\documentclass[letterpaper]{article}
\usepackage{aaai}
\usepackage{times}
\usepackage{helvet}
\usepackage{courier}
\usepackage{bbm}
\usepackage{graphicx}
\usepackage{algorithmic}
\usepackage{algorithm}
\usepackage{xcolor}
\usepackage{amsmath}
 
\begin{document}

\title{Robust Deep Learning with Active Noise Cancellation for Spatial Computing}
\author{Li Chen,\textsuperscript{1}
David Yang,\textsuperscript{2}
Purvi Goel,\textsuperscript{3}
Ilknur Kabul\textsuperscript{4}\\
{Facebook}\\

\textsuperscript{1}lichen66@fb.com,
\textsuperscript{2}dzyang@fb.com,
\textsuperscript{3}purvigoel@fb.com,
\textsuperscript{4}ilknurkabul@fb.com
}
\maketitle
\begin{abstract}
\begin{quote}

This paper proposes CANC, a \textbf{C}o-teaching \textbf{A}ctive \textbf{N}oise \textbf{C}ancellation method, applied in spatial computing to address deep learning trained with extreme noisy labels. Deep learning algorithms have been successful in spatial computing for land or building footprint recognition. However a lot of noise exists in ground truth labels due to how labels are collected in spatial computing and satellite imagery. Existing methods to deal with extreme label noise conduct clean sample selection and do not utilize the remaining samples. Such techniques can be wasteful due to the cost of data retrieval. Our proposed CANC algorithm not only conserves high-cost training samples but also provides active label correction to better improve robust deep learning with extreme noisy labels. We demonstrate the effectiveness of CANC for building footprint recognition for spatial computing. 

\end{quote}
\end{abstract}

\section{Introduction}
Deep learning has shown success in a variety of computer vision applications. In spatial computing, deep learning has demonstrated superior performance in road or building footprint recognition as well as population density estimation. However one undesired property of deep learning can hinder its performance badly: in the presence of noisy labels, which can arise due to measurement errors, crowdsourcing, insufficient expertise and so on, deep learning tends to have poor generalization performance on test data. The neural networks fit to correctly-labeled training samples at earlier epochs, but eventually overfit to samples with noisy labels, leading to poor classification performance on the test set. This is the so-called memorization effect of deep learning in the presence of noisy labels. \cite{han2018co,arpit2017closer,algan2020label}. 

\begin{figure}[h]
\includegraphics[width=8cm]{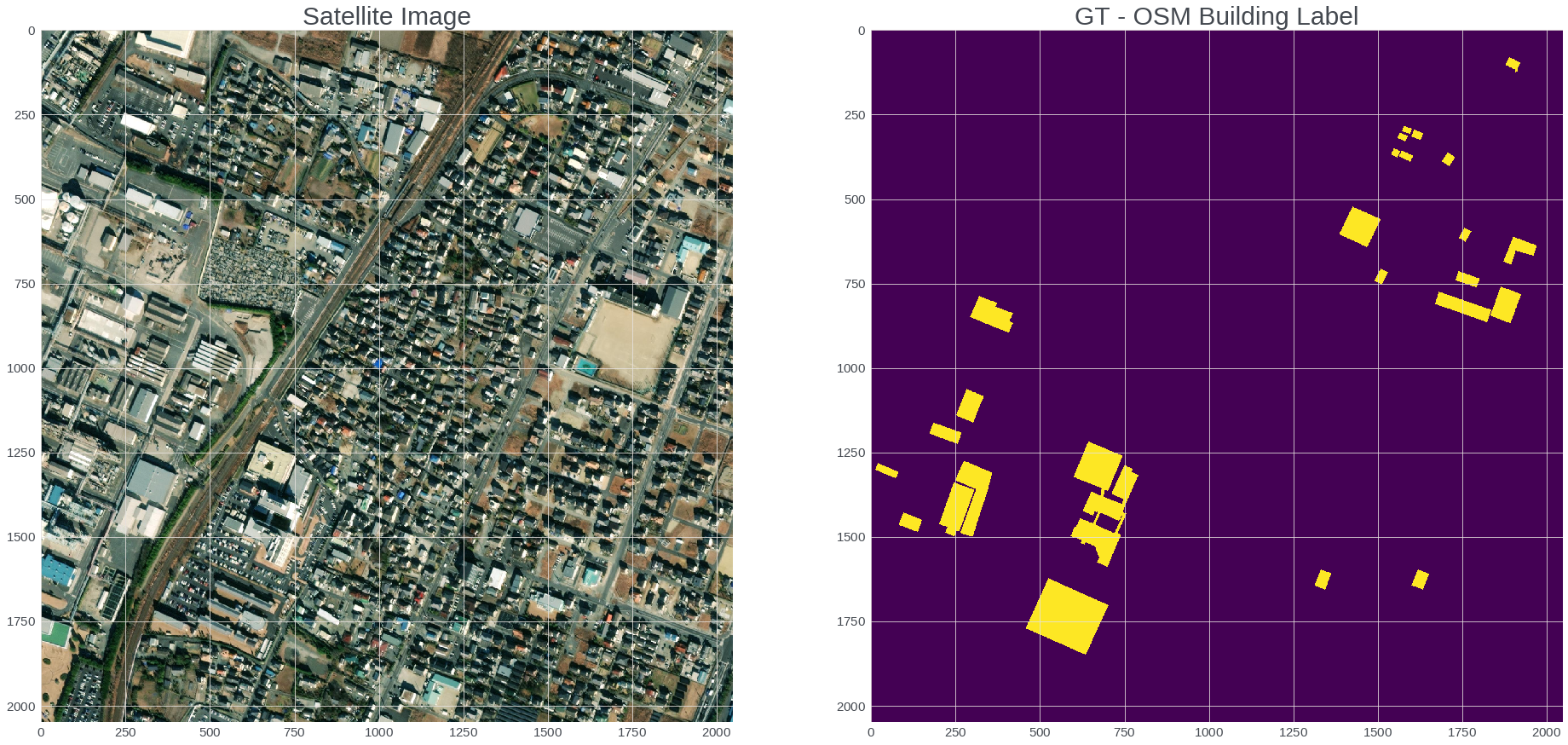}
\caption{Left: satellite image shows a large number of buildings in the region. Right: Labels from Open Street Map (OSM). OSM only has labels for a few of them. A lot of building footprints are missing due to label crowdsourcing. This is considered as extreme noise with the majority of labels being sparse or missing.}\label{fig:noisy_sat}
\end{figure}

In spatial computing, noisy labels, largely inevitable, come from two major sources. The first source is crowdsourcing, a collaborative initiative to create a free map with participants contribute to labeling the map. One famous and widely-used database for spatial computing is Open Street Map (OSM) \cite{OpenStreetMap}, a free and editable map of the whole world that is being built by volunteers largely from scratch and released with an open-content license. The labels can be contributed by mappers either on foot or bicycle or in a car or boat. If a location is not widely explored by the mappers, it is likely the labels are noted wrongly. This type of noise can be extreme, meaning the majority of the labels are either sparse or wrong, as seen in Figure \ref{fig:noisy_sat}. The second major source of noisy labels in spatial computing is measurement error. Imagery providers usually do a good job at geo-referencing their imagery, but occasionally the images can be out of position by a few meters or more. Particularly in hilly or mountainous areas, it is difficult to stretch a flat image over an area of the Earth with many contours. Then even with professional annotators, the resulting ground truth can be offset due to the imagery misalignment. 

Co-teaching, proposed in \cite{han2018co}, has demonstrated efficacy to improve deep learning robustness in the presence of extreme noisy labels. It involves training two neural networks simultaneously such that each network teaches the other for clean sample selection. As a clean data selection method, while co-teaching outperforms a variety of noisy label learning methods such as MentorNet \cite{jiang2018mentornet}, Decoupling \cite{malach2017decoupling}, S-net \cite{goldberger2016training}, Bootstrap \cite{reed2014training}, it can filter a large number of training samples via a parameter called remember rate. It is trained only on a smaller portion of the data samples as epochs grow to unsure robust learning. However, such procedure may not take full advantage of high-cost commercial satellite imageries in spatial computing. 
In addition, co-teaching can help to find the data with less noise and use them more often during training, but it could not actively correct the noise. In the case of spatial computing where there is not enough data with low noise , often estimated to be less than $1\%$,  co-teaching might not achieve good performance due to the lack of data with low noise.

Hence, in order to address learning with extreme noisy labels and conserving valuable data samples in spatial computing, we propose CANC, \textbf{C}o-teaching \textbf{A}ctive \textbf{N}oise \textbf{C}ancellation, a training paradigm to not only perform clean sample filtering but also actively correct unreliable labels. Our method makes better use of training dataset compared to co-teaching. Especially under higher noise strengths, CANC outperforms co-teaching because CANC's active label correction capability enables utilizing more training data to train a resilient deep learning model. We also notice, however, when the noise strength is less, CANC does not outperform co-teaching. In this paper, we demonstrate the effectiveness of CANC in building footprint recognition under extreme noise and evaluate the performance using accuracy, precision, recall, F1 score and super-pixel intersection over union.

\section{Related Work}

Deep learning has the tendency to overfit data and even memorize completely random noise. Such undesired property of deep learning inspired several research directions in robust deep learning with noisy labels. Two main categories of methods exist, namely noise-model-based and noise-model-free methods. 
The task of noise-model-based methods is to identify the best estimator via minimizing a risk function to describe the effects of noisy samples.  Existing methods include noise channel estimation, label noise cleansing, dataset pruning and sample importance weighting. The authors in \cite{northcutt2017learning} estimate the noise rates according to the sizes of a confidently clean and a noisy subset.   \cite{wu2018light}  uses transfer learning trained on a clean dataset and fine-tunes it on noisy dataset for relabeling and then the network is retrained on relabeled data to re-sample to dataset to construct a final clean dataset. The authors in \cite{jiang2018hyperspectral} propose to randomly split dataset to labeled and unlabeled subgroups. Then, labels are propagated to unlabeled data using similarity index among instances. 
\cite{wang2018iterative}  uses the Siamese network to detect noisy labels by learning discriminating features to apart clean and noisy data. Noisy samples are detected and pulled from clean samples. Then, each iteration weighting factor is recalculated for noisy samples, and the base classifier is trained on whole dataset. 

On the other hand, the task of noise-model free methods is to design robust algorithms rather than modeling the noise. In that sense, label noise is not decoupled from classification. Non-convex loss function is more noise tolerant than convex losses as studied in  \cite{manwani2013noise}, \cite{ghosh2015making}, \cite{charoenphakdee2019symmetric}. \cite{xu2019l_dmi} proposed to use information-theoretic loss.  \cite{van2015learning} proposed to use classification-calibrated loss function. Regularization methods such as dropout, adversarial training, label smoothing are useful for addressing noisy label problems. Boosting algorithms such as BrownBoost and LogitBoost are shown to be more robust to noisy labels \cite{mcdonald2003empirical}. 

Building footprints are critical for many applications, such as mapping, city and government planning, population density estimation \cite{Andi2017PopDen}, and disaster estimation. Manual building detection for the whole world takes a long time and requires a huge amount of human resources such that it might be infeasible. Recently, deep learning has been applied to extract map features from satellite images, such as roads and buildings \cite{Kaiser2017_noisy} \cite{Derrick2019_noisy} \cite{Anil2019} \cite{chawda2018} \cite{Pan2019}. Among these works, \cite{Anil2019}, \cite{chawda2018}, and \cite{Pan2019} used professionally labeled training dataset with little noise, such as DeepGlobe dataset \cite{Deepglobe2018}, SpaceNet dataset \cite{spacenet2019}, and Massachusetts Building dataset \cite{Minh2013}. As these datasets are limited to a few selected areas, the models trained on them might not generalize well across the globe. In contrast, \cite{Kaiser2017_noisy} and \cite{Derrick2019_noisy} used the crowd-sourced map dataset OSM for training, but the OSM dataset has high noise-to-signal ratio.
\section{Framework and Methodology}
\begin{figure}[h]
\includegraphics[width=8cm, height=4cm]{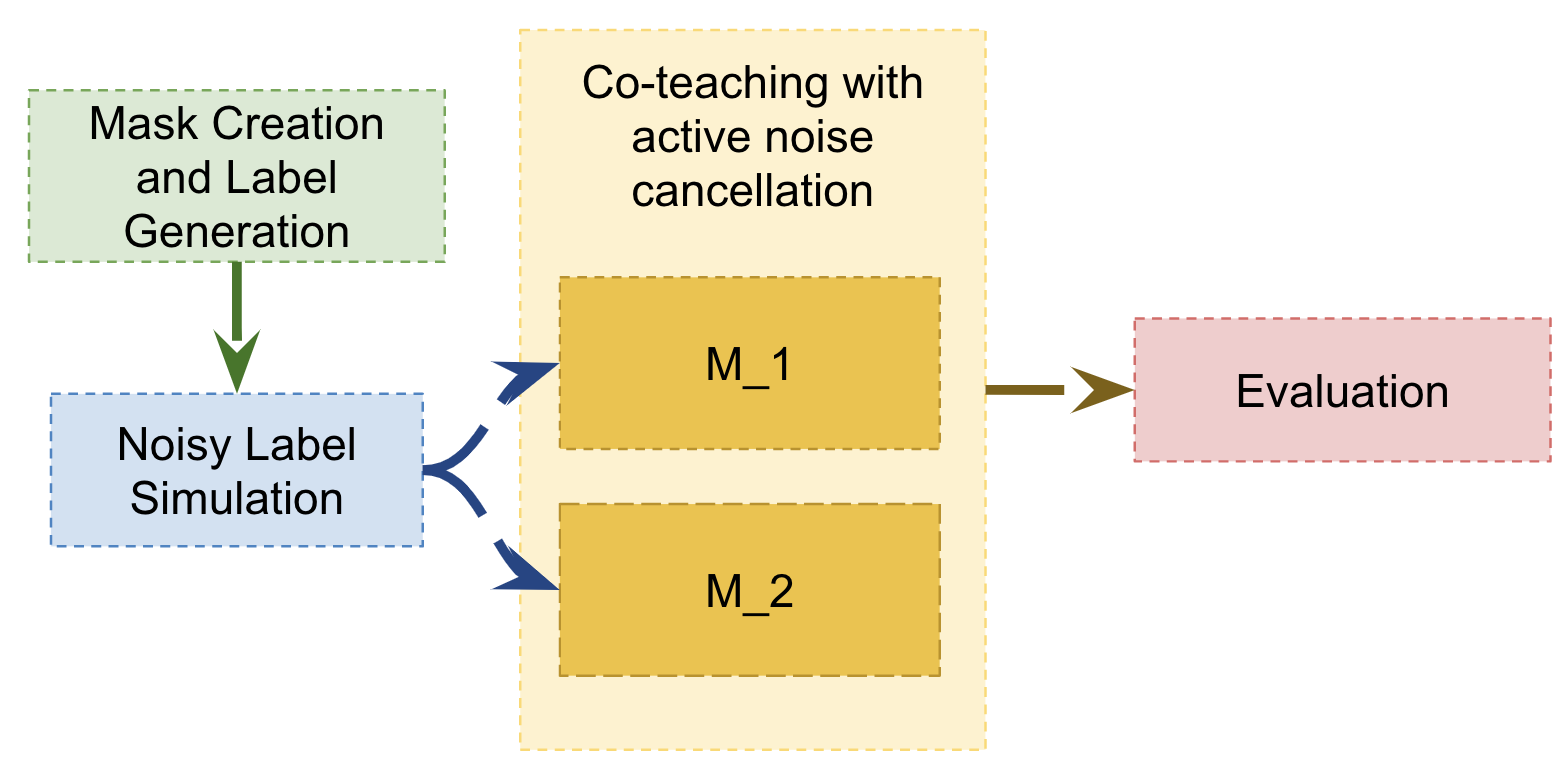}
\caption{Overview of our proposed framework. We consider a sub-task called mask classification task, where the masks and associated labels are created from the original satellite imagery. Then we perturb the labels via a noise transition matrix. CANC is trained on the noisy data to produce a robust deep learning algorithm. Finally we evaluate the performance using accuracy, precision, recall, F1 and super-pixel intersection over union. }\label{fig:overview}
\end{figure}

Our proposed framework, as seen in Figure \ref{fig:overview}, consists of the following four main steps: 1. Mask creation and label generation, 2. Noisy label simulation 3. CANC: co-teaching with active noise cancellation 4. Evaluation.
We describe each component in detail.

\subsection{Mask Creation}
Given a high-resolution satellite image, we generate masks of size $m \times m \times 3$ by cropping the original image of size $n \times n \times 3$, where $m$ is divisible by $n$. Hence each satellite image is represented by $[\frac{n}{m}]^2$ masks. 

Similarly on each ground truth satellite image, we create grey-scale masks with size $m \times m \times 1$ to represent the annotated area. 
The label for each mask is computed as follows:
\begin{equation}
    l = \mathbbm{1}_{\{\frac{S_1}{m^2} \geq \tau\}},
\end{equation}
where $\mathbbm{1}$ is the identity function, $S_1$ is the sum of the pixels that belong to building class, and $\tau$ is a threshold set to represent the percentage of pixels needed in one mask for a professional annotator to recognize a building in it. $\tau = 0.5\% \text{ or } 1\%$ is usually a reasonable value.  

\subsection{Noise Simulation}
Our framework then simulates noise to ground truth so we can use as a baseline to assess the effectiveness of the framework. T
We consider two types of noise simulation schemes: symmetric and anti-symmetric. Both of these schemes require manipulation of a square noise transition matrix used to choose which data samples to mislabel.

In the symmetric scheme, our transition matrix has a diagonal entries of (1 - $\epsilon$), where $\epsilon$ is a predefined noise ratio indicating the percentage of data samples to mislabel. Everywhere else but the diagonal, the matrix has entries of  $(1 - \epsilon) / (n-1) $ where n is the number of classes. Intuitively, this transition matrix means that data samples can be mislabeled as any other class with equal probability.

In the anti-symmetric scheme, we use a transition matrix
\begin{equation}
  \begin{bmatrix}
    1 - \epsilon & 0 \\
    \epsilon & 1 
  \end{bmatrix}
\end{equation}
 
 Again, $\epsilon$ is the chosen noise ratio. Intuitively, this matrix suggests that all of class 1 will stay the same, and class 0 is flipped to class 1 with a probability equal to $(1 -  \epsilon)$. 


\subsection{Co-teaching with Active Noise Cancellation} \label{subsec:canc}
We present our proposed Co-teaching with Active Noise Cancellation (CANC) as seen in Algorithm \ref{alg:canc} in this subsection. We leverage co-teaching \cite{han2018co}, which is a training paradigm to address noisy label learning robustness. In co-teaching training scheme, there are two neural networks where two networks are simultaneously trained and taught by each other given every mini batch. It essentially consists of three steps: i) each neural network feeds forward all data and performs clean sample selection based on losses, ii) two networks exchange what data shall be used for training, iii) each network trains on the data selected by the other teacher and updates itself. Co-teaching has shown to be robust especially under the extreme noisy label scenario.

Our goal is to not only train a robust deep learning algorithm in the presence of noisy labels, but also actively correct the noise during the training process without hurting classification performance. To make better use of the data, instead of forgetting the remaining samples in the batch as in co-teaching, we consider swapping the labels of the samples that are most likely to have noisy labels. Since our classification problem is binary, the swapping process becomes a binary decision. Our method inherits co-teaching such that the loss is sorted in each mini-batch by the first network which sends the top samples with least amount of loss to the other network. Additionally our method would select the bottom samples with the highest amount of loss, correct their labels and combine with the first subset to update the parameters of the second network. The same applies for updating the parameters of the first network.  The algorithm is stated in Algorithm \ref{alg:canc}.

\begin{algorithm}
\label{alg:canc}
\caption{Co-teaching with active noise cancellation (CANC)}
\begin{algorithmic}

\STATE \textbf{Goal}: Robust training on extreme noisy labels with active noise correction
\STATE \textbf{Input}: $w_{M_1}$, $w_{M_2}$, learning rate $\eta$, epoch $T_k$, $T_{\max}$, iteration $N_{\max}$, remember rate $R(T)$, swap rate $S$;
\FOR {$T = 1,2, ..., T_{\max}$}
    \FOR {$N=1,...,N_{\max}$}
    \STATE Step 1: Feed mini-batch $\mathcal{D}$ to both $M_1$ and $M_2$
    \STATE Step 2: Obtain \\$\hat{D}^{clean}_{M_1} = \arg\min_{D^{\prime}:|D`|\geq R(T)|D^{\prime}|}l(M_1, D^{\prime})$
    \STATE Step 2: Obtain \\$\hat{D}^{swap}_{M_1} = \arg\max_{D^{\prime}:|D`|\geq S|D^{\prime}|}l(M_1, D^{\prime})$
    \STATE Step 3: Define $\hat{D}_{M_1}:= \hat{D}^{clean}_{M_1} \cup \hat{D}^{swap}_{M_1}$ and update $w_{M_2} \leftarrow w_{M_2} - \eta \nabla (M_2, D_{M_1})$
    \STATE Step 4: Obtain \\$\hat{D}^{clean}_{M_2} = \arg\min_{D^{\prime}:|D`|\geq R(T)|D^{\prime}|}l(M_2, D^{\prime})$
    \STATE Step 5: Obtain \\$\hat{D}^{swap}_{M_2} = \arg\max_{D^{\prime}:|D`|\geq S|D^{\prime}|}l(M_2, D^{\prime})$
    \STATE Step 6: Define $\hat{D}_{M_2}:= \hat{D}^{clean}_{M_2} \cup \hat{D}^{swap}_{M_2}$ and update $w_{M_1} \leftarrow w_{M_1} - \eta \nabla (M_1, D_{M_2})$
    \ENDFOR
\STATE Step 7: Update $R(T) = 1 - \min\{\frac{T}{T_k}\tau, \tau \}$
\ENDFOR
\STATE Output: $w_{M_1}$ and $w_{M_2}$
\end{algorithmic} \label{alg:canc}
\end{algorithm}

\subsection{Evaluation}
We evaluate the accuracy, precision, recall, $F_1$ score and super-pixel intersection over union (SP-IOU) to assess the performance of our algorithm on spatial computing, where each super-pixel corresponds to the generated mask. We define a smoothed SP-IoU as the following:
\begin{equation}
    \mbox{SP-IoU} = \frac{intersection + smooth}{union + smooth} ,
\end{equation}
where $intersection = \|\{ y_i == \hat{y}_i\}\|_{\{ 1\leq i \leq N\}}$, $union = \|\{ y_i \}\|_{\{ 1\leq i \leq N\}} + \|\{\hat{y}_i\}\|_{\{ 1\leq i \leq N\}} - intersection$, $y_i$ and $\hat{y}_i$ are the label and predicted label of mask $i$ respectively and $\| \cdot \|$ denotes the cardinality. We set $smooth$ parameter to be 1.

\section{Data Description}

The inputs 
are high resolution satellite images from Maxar. Each pixel in the images represents an area of 50cm by 50cm on the ground. The satellite images are tiled at zoom level 15 with the size of 2048 by 2048, so each satellite image represents about 1km by 1km on the ground. The building labels of the training dataset were collected from OpenStreetMap (OSM). OpenStreetMap is a free, editable map of the whole world that is being built by volunteers largely from scratch and released with an open-content license. 
The building labels, in the form of polygons, are rasterized to create the training label images. 
\section{Experiments}

\subsection{Data Preparation}
We obtain 81 satellite imageries with professional annotation, which means we have minimum noisy labels. We put 50 satellite images as training, 15 as validation and 15 as testing.
The original satellite imageries are of size $8192 \times 8192 \times 3$. We set the mask size to be $32 \times 32$. Then each satellite image is segmented into $256\times 256$ masks. We further set the threshold $\tau = 1\%$ since with approximately 8 pixels, the annotators can recognize this part of a building.  Hence after preprocessing, we obtain 3.34 million masks for training, 983k masks for validation and 15 images for another testing. 

We report accuracy, precision, recall and $F_1$ score on the validation masks and report the super-pixel IOU on the 15 test images. 
\subsection{Label Noise Simulation}
We use several methods to apply label noise to our clean training labels. Because of our data preparation, in which we cropped and labelled images with either a 0, for no buildings, or a 1, for containing buildings, we've reparameterized the DNN's decision into a binary classification problem. As follows, label noise can be added by perturbing these 0/1 labels rather than the position or presence of building polygons themselves.

\textit{Symmetric Noise Simulation}. In the symmetric noise strategy, we flip one label to another based on a transition probability matrix T. The length and width is equal to the number of classes; in this binary classification case, the matrix dimensions are 2 x 2. The diagonal entries of the transition matix are (1 - N), where N is some specified noise ratio between 0 and 1, which indicates a rough percentage of labels to be perturbed. We distribute the remaining probability equally per-row as: (N) / c, where c is the number of classes. An example transition matrix, with noise ratio. An example matrix, calculated with a noise ratio of 0.35, is shown below:
\begin{equation}
  \begin{bmatrix}
    0.65 & 0.35 \\
    0.35 & 0.65 
  \end{bmatrix}
\end{equation}

\textit{Antisymmetric Noise Simulation}. In the antisymmetric noise strategy, we add noise to only single class. As before, we use a 2x2 noise transition matrix. This time, the matrix flips the class with probability equal to (1 - noise ratio), using a matrix such as the following for a noise ratio of 0.35.

\begin{equation}
  \begin{bmatrix}
    0.65 & 0 \\
    0.35 & 1 
  \end{bmatrix}
\end{equation}

We perform our experiments with varying noise strengths: 0.15, 0.35, 0.45, and 0.55.

\subsection{Results}
We apply CANC on the noisy training masks and compare the performance with deep learning directly training with noisy labels and co-teaching without active noise cancellation. We consider both transfer learning via ResNet18 \cite{he2016deep} and training from scratch utilizing a 9-layer CNN with LReLU architecture as in \cite{han2018co,laine2016temporal} on the dataset, and do not observe significant performance difference between training from scratch and transfer learning in our use case. Within the 3.34 million training masks, we use 80\% to update the weights and 20\% as validation for model selection at the best validation accuracy. 
On a second validation set consisting 983k masks, we evaluate the model performance and report accuracy, precision, recall and $F_1$ score. Finally we evaluate the super-pixel IoU per satellite image from the test set which contains another 15 out-of-sample satellite images. 

The performance of CANC, co-teaching and vanilla deep learning is shown in Table \ref{tab:validation_perf}.  The clean baseline indicates the model's performance on validation set when the model was trained with training data with clean labels. The noisy baseline indicates the model's performance on validation set when the model was trained with training data with noisy labels generated by different noise types and at different noise strength. For lower noise ratio at 0.15, CANC does not outperform co-teaching. As the noise ratio is increased higher to 0.45 and 0.55, CANC improves network performance compared with directly training on noisy labels and outperforms co0teaching in accuracy, precision, recall and F1 scores.

\begin{table*}
    \centering
    \begin{tabular}{|l|c|c|c|c|c|c|}
    \hline
        Scenario & Noise strength &Noise type &Accuracy& Precision& Recall& $F_1$\\ \hline
         Clean baseline& Clean &Clean &0.9350 &0.9844&0.9324&0.9577\\
         Noisy baseline& 0.15 & Anti-symmetric& 0.8226 &0.9915 & 0.8089& 0.8910\\
         CANC &0.15 & Anti-symmetric& 0.6899  & 0.8702 & 0.7684 & 0.8161  \\
         Coteaching &0.15 & Anti-symmetric& 0.9209  & 0.9686 &0.9422  & 0.9552 \\
         Noisy baseline& 0.35& Anti-symmetric &0.1164  & 0.9736  & 0.0142 & 0.0280 \\
         Noisy baseline& 0.35 &Symmetric &0.3684&nan&0&nan\\
         Coteaching &0.35 &Symmetric &0.9290&0.9514&0.9739&0.9625\\
         Noisy baseline& 0.45 &Anti-symmetric &0.1040  &  nan& 0 & nan\\
         CANC &0.45&Anti-symmetric &  0.8870& 0.9833& 0.8889 & 0.9337 \\
         Coteaching &0.45 &Anti-symmetric &  0.8649& 0.9827& 0.8643& 0.9197\\
         Noisy baseline& 0.55 &Anti-symmetric &0.1040   &  nan& 0 & nan\\\
         CANC &0.55 & Anti-symmetric & 0.8944   & 0.8958  &  0.9982&  0.9442\\
        Noisy baseline& 0.55 &Symmetric&0.1040 &0.1176 & 0 &0\\
        CANC &0.55 & Symmetric &0.9190 &0.9291 & 0.9847&0.9561 \\
        CANC (S = 1-R) &0.55 &Symmetric& 0.1637&0.9866& 0.0675&0.1263\\
        Coteaching &0.55 &Symmetric& 0.1040 &nan&0&nan\\

\hline
    \end{tabular}
    \caption{Performance on validation set when training with various noise strengths and noise types.}
    \label{tab:validation_perf}
\end{table*}

\begin{figure}[h]
\centering
\includegraphics[width=9cm, height=5cm]{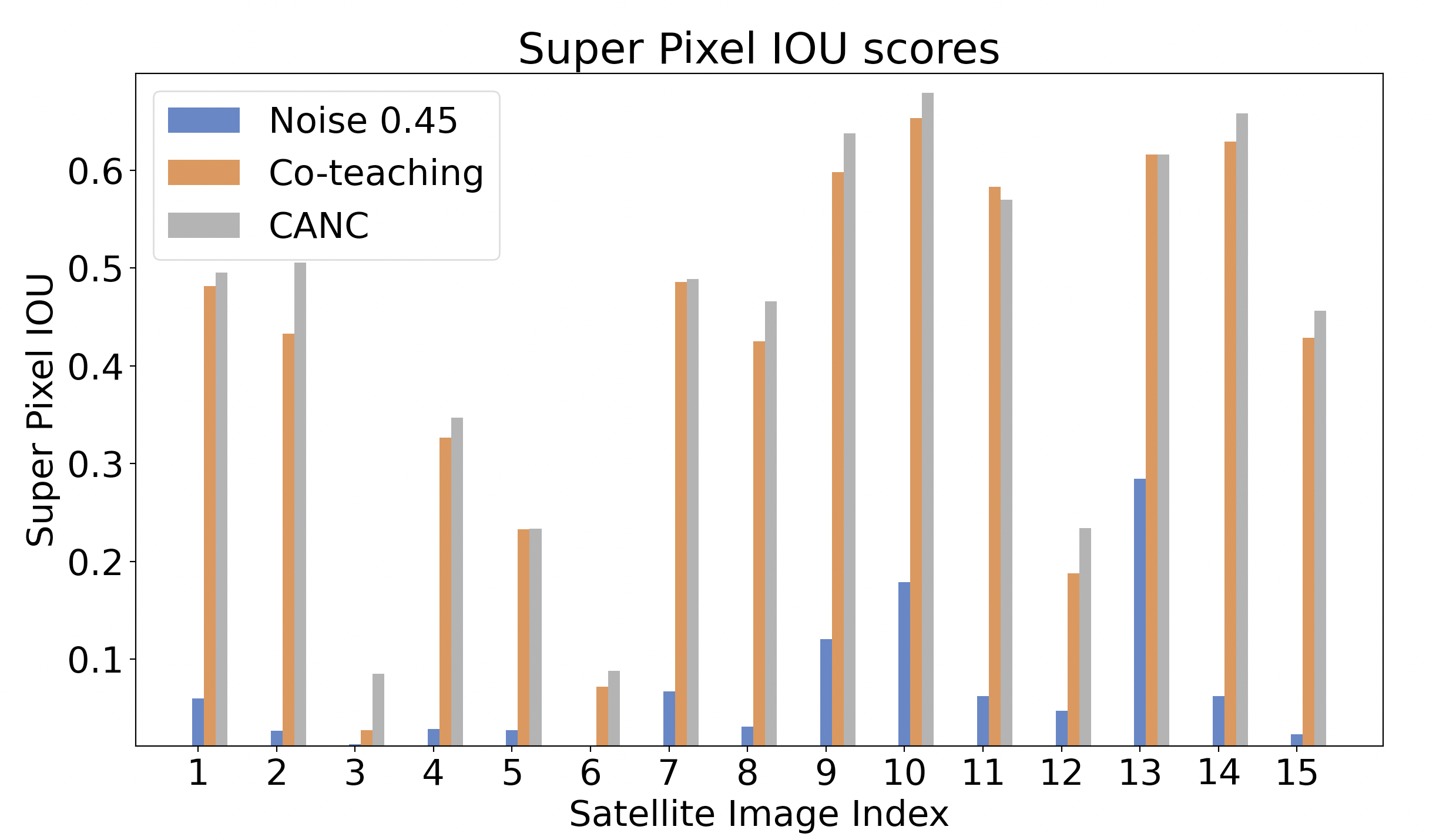}
\caption{We calculate the super-pixel IoU under noise ratio of 0.45 for anti-symmetric noise for 15 test images. CANC with improves the SP-IoU compared with deep learning without addressing noisy labels. It also outperforms co-teaching on most of the satellite images.}\label{fig:0.45spiou}
\end{figure}

\begin{figure}[h]
\centering
\includegraphics[width=9cm, height=5cm]{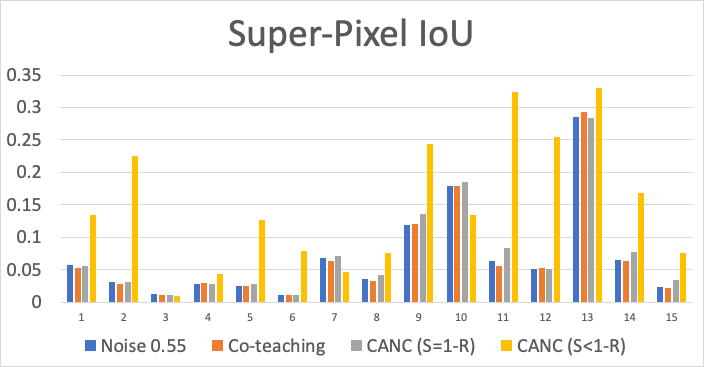}
\caption{We calculate the super-pixel IoU under noise ratio of 0.55 for symmetric noise for 15 test images. In this experiment, we additionally considered the case when the swap rate is the same as the initial forget rate in co-teaching, which we denote as CANC with $S = 1-R$. CANC with $S < 1-R$ improves the SP-IoU the most. Both CANC methods can improve super-pixel IOU when training with noisy labels.}\label{fig:0.55spiou}
\end{figure}

On the 15 images in test set, we plot the super-pixel IoUs and compare this metric as seen in Figure \ref{fig:0.45spiou} and \ref{fig:0.55spiou} under two noise types and strengths respectively. 

\subsection{Inspection}
We examine which images benefit the most from CANC in terms of SP-IoU when the model has been learned with symmetric noise at 0.55 noise strength. The improved imagery has very different geo-condition compared with the worst improved imagery. The best improved image has buildings in a much smaller location. 
We think the performance improvement is due to the model being fed by corrected information about buildings in such geo-condition. However we acknowledge that the false positive rate also increases. We plan to do more investigation on interpreting the results.

\begin{figure}[h]
\centering
\includegraphics[width=4cm, height=4cm]{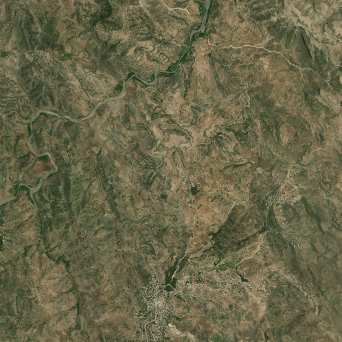}
\includegraphics[width=4cm, height=4cm]{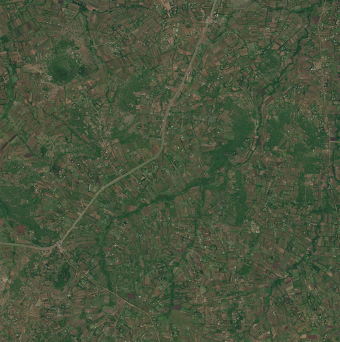}

\caption{(Left) The satellite image with best SP-IoU improvement (7.35x better) with CANC under noise ratio 0.55 of symmetric noise. (Right) The satellite image with worst SP-IoU improvement (0.67x better) with CANC under noise ratio 0.55 of symmetric noise. We think the performance improvement is due to the model being fed by corrected information about buildings in such geo-condition. However we acknowledge that the false positive rate also increases.
}\label{fig:0.55best_improvement}
\end{figure}
\section{Conclusion}

Having high quality training data is an important part for the success of the training methods in machine learning. For classification problems, label quality plays an important role for the data quality. Obtaining high quality ground truth labels is a very challenging problem due to the scarcity of labelling resources, inconsistencies between the labellers, as well as the measurement errors that happens during the label collection process. In this paper, by using CANC we are able to mitigate some of these issues and improve the predictive accuracy of our models. 
 We validated the proposed method on real data and obtained promising results. We showed that the proposed technique has advantages over co-teaching in that it corrects some of the labelling errors in the training data during the training process rather than removing them. We also acknowledge that under certain noise strength, co-teaching demonstrates more benefits over CANC.
 
 \section{Future Work}
 In our noise simulation scheme, we applied noise directly to the mask labels. We are working on a noise simulation process directly applied on the building polygons. 
In this simulation process, we artificially add buildings to images, where the buildings are selected from aggregated statistical analyses on all the real-world buildings. Within a single given satellite image, we collect all the ground-truth buildings shapes and sort them, based on their image-space polygonal area, into a histogram. We normalize the histogram to convert it into a distribution of per-image building areas and randomly sample a bucket from the distribution, then sample a building shape from that bucket. The building is placed at a randomly chosen coordinate within the image. We continue this process of sampling-and-adding until the ratio of pixels labeled as buildings to the total number of pixels is roughly equal to a predefined threshold. The challenge to make such simulation process realistic is due to building boundaries. An added polygon shall not have boundaries to overlap with existing polygons. While we can employ an acceptance-rejection method, it significantly increases the noise simulation run time. We plan to investigate an efficient method to perturb labels directly in the building polygon-level.

In this use case, we mainly focus on binary classification problem, in which we rely on the loss in the mini-batch to flip the training data labels. For future work, we would also like to extend it to multi-family classification. We plan to consider a weighted decision to flip the labels based on the predicted probabilities. 

Our active noise cancellation process via a swapping procedure is not sophisticated. We would like to investigate in the future is incorporating uncertainty estimation to CANC to improve the robustness of the models to label noise. We would like to investigate both epistemic and aleatory uncertainty techniques in this research direction.

\bibliographystyle{aaai}
\bibliography{references}
\end{document}